\pdfoutput=1
\documentclass[11pt]{article}
\usepackage[final]{acl}
\usepackage{times}
\usepackage{latexsym}
\usepackage[normalem]{ulem}
\usepackage{xcolor}%
\usepackage{tcolorbox} 
\tcbuselibrary{breakable}
\usepackage{enumitem}
\usepackage{xspace}
\usepackage{textcomp}%
\usepackage{hyperref}
\usepackage{url}
\usepackage{amssymb}
\usepackage{amsmath}
\usepackage{bm}
\usepackage[T1]{fontenc}
\usepackage[utf8]{inputenc}
\usepackage{microtype}
\usepackage{inconsolata}
\usepackage{graphicx}
\usepackage{booktabs}
\usepackage{array}
\usepackage{geometry}
\usepackage{caption}
\usepackage{multirow}
\usepackage{makecell}

\usepackage{lipsum}
\title{SeMob: Semantic Synthesis for Dynamic Urban Mobility Prediction}

\author{
  Runfei Chen\textsuperscript{1} \and
  Shuyang Jiang\textsuperscript{3} \and
  Wei Huang\textsuperscript{1,2,4}\thanks{Corresponding Author.} \\
  \textsuperscript{1}Urban Mobility Institute, Tongji University, China \\
  \textsuperscript{2}College of Surveying and Geo-informatics, Tongji University, China\\
  \textsuperscript{3}College of Computer Science and Artificial Intelligence, Fudan University, China\\
  \textsuperscript{4}Department of Civil Engineering, Toronto Metropolitan University, Canada \\
  \texttt{\{runfeichen, wei\_huang\}@tongji.edu.cn, shuyangjiang23@m.fudan.edu.cn}
}

\begin{document}
\maketitle
\begin{abstract}
Human mobility prediction is vital for urban services, but often fails to account for abrupt changes from external events. Existing spatiotemporal models struggle to leverage textual descriptions detailing these events. We propose SeMob, an LLM-powered \underline{\textbf{sem}}antic synthesis pipeline for dynamic \underline{\textbf{mob}}ility prediction. 
Specifically, SeMob employs a multi-agent framework where LLM-based agents automatically extract and reason about spatiotemporally related text from complex online texts. Fine-grained relevant contexts are then incorporated with spatiotemporal data through our proposed innovative progressive fusion architecture. The rich pre-trained event prior contributes enriched insights about event-driven prediction, and hence results in a more aligned forecasting model. 
Evaluated on a dataset constructed through our pipeline, SeMob achieves maximal reductions of 13.92\% in MAE and 11.12\% in RMSE compared to the spatiotemporal model. Notably, the framework exhibits pronounced superiority especially within spatiotemporal regions close to an event's location and time of occurrence.
\end{abstract}

\section{Introduction}
\begin{figure}[tbp]
  \centering 
  \includegraphics[width=\linewidth]{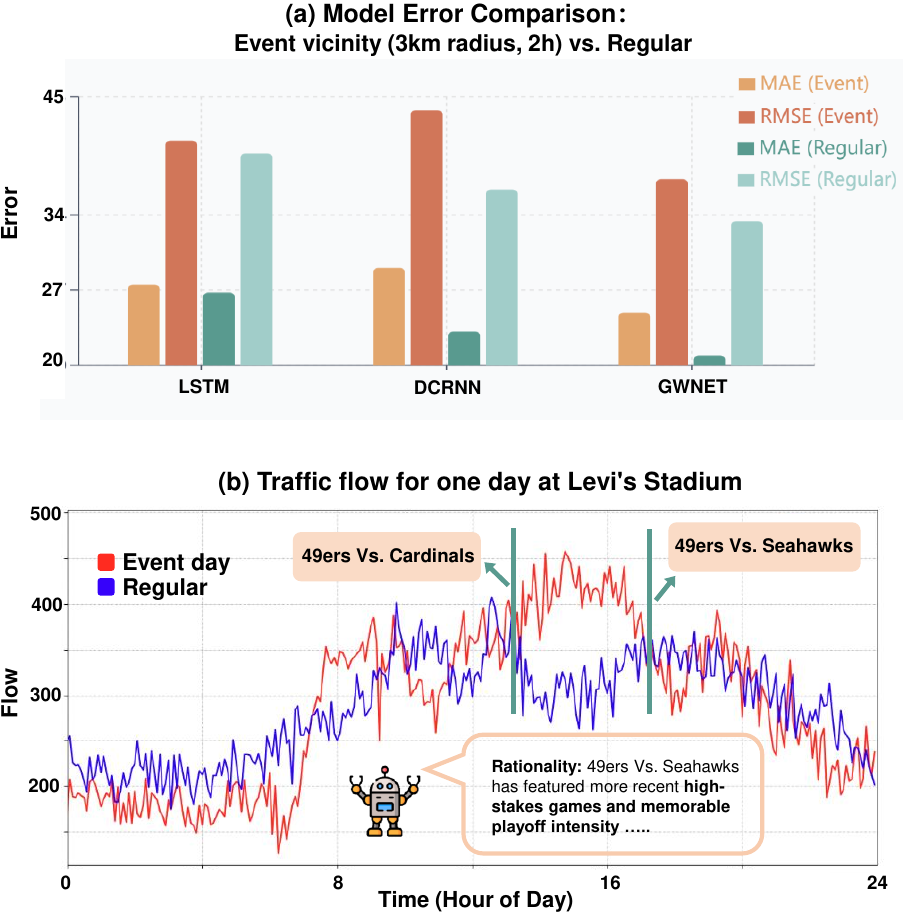}
  \caption{Challenges in event-based mobility prediction. (a) Spatiotemporal models, effective during regular days, can exhibit higher errors than the simpler LSTM. (b) Complex event semantics drive pattern variance, even in similar event types.}
  \label{fig:intro}
\end{figure}

Human mobility prediction is an important part of urban services optimization, including intelligent routing and dynamic traffic management~\citep{li2024optimization,moon2025traffic}. While established methods capture routine mobility patterns by historical data and predefined graph structures~\citep{yin2021deep,liu2023largest,wang2024stone}, they struggle to interpret and adapt to abrupt changes caused by various external events, as shown in Figure~\ref{fig:intro}(a). The unique nature of urban spatiotemporal dynamics stems from the complexity of human motivations and the diversity of events driving mobility~\citep{gong2024mobility,han2025adapting,bontorin2025mixing}, making their underlying semantics difficult for standard models to capture. Information detailing driving events is largely conveyed through textual descriptions, commonly originating from sources such as official websites and social media~\citep{mihalcea2024developments,pappalardo2023future}. A significant gap therefore exists in leveraging the understanding embedded in the language to improve the responsiveness and accuracy of mobility prediction.

\begin{figure*}
    \centering
    \includegraphics[width=\linewidth]{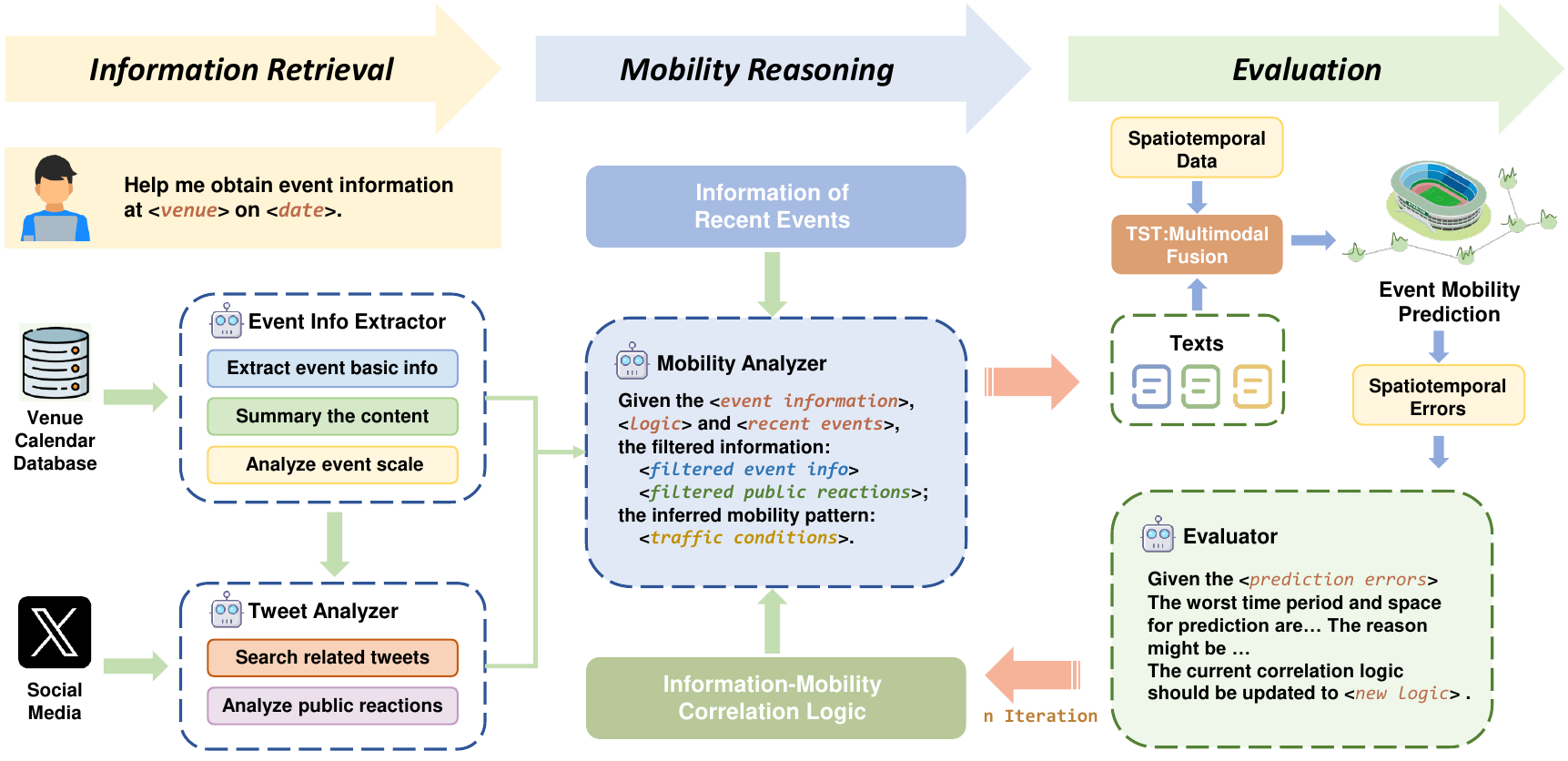}
    \caption{SeMob framework. Multi-agents iteratively collect spatiotemporally relevant texts for TST multimodal prediction and refine the text filtering logic using prediction feedback.}
    \label{fig:multi-agent}
\end{figure*}
Approaches transforming text information to numerical data or discrete categories often fail to capture context-dependent variations~\citep{liang2024exploring}. An example in Figure~\ref{fig:intro}(b) demonstrates superficially similar events can trigger markedly different mobility responses. Descriptive texts associated with event backgrounds offer a promising avenue for unraveling the intricate mechanism of human behaviors and societal changes. Exploring the integration of qualitative insights from such descriptions may enable models to capture complex, non-linear influences missed by purely quantitative approaches. However, harnessing textual data in this domain confronts notable challenges. 

A primary difficulty involves the effective extraction and representation of pertinent semantics from textual sources. Mobility-related events suffer from a notable scarcity of relevant textual data, limiting the depth of contextual analysis~\citep{wang2024news,han2024event}. Furthermore, existing models that combine textual and temporal data~\citep{liu2024autotimes,cao2024tempo,hu2025context} lack consideration of spatial dimension. Current fusion approaches typically align temporal sequences with the corresponding series descriptions but fail to incorporate external semantic information meaningfully. Creating effective methods to fuse unstructured textual representations with continuously updated structured spatiotemporal data remains a key research objective~\citep{zou2025deep}.

We introduce a novel pipeline SeMob to integrate event insights into human spatiotemporal mobility prediction. The multi-agent framework in SeMob extracts and reason about relevant information from online sources. This task extends beyond simple keyword matching; it requires a deep understanding of how textual elements relate to spatiotemporal forecasts and calls for advanced analytical reasoning. These agents simulate traffic analysis workflows to identify event-related text. The extracted text is then paired with the corresponding spatiotemporal data to create context-aware mobility datasets that improve prediction accuracy. Furthermore, the LLM agent drives an iterative refinement process for text extraction. An agent in the workflow compares model predictions with ground truth mobility flow, thereby uncovering crucial, previously overlooked logical connections. Through this iterative analysis of unstructured text, the agent further identifies patterns linking textual cues to prediction discrepancies and provides valuable, hard-to-acquire textual insights.

Fusing extracted textual information with structured spatiotemporal data is another key challenge in enhancing mobility forecasting. In SeMob, we propose a progressive fusion method that combines contextual insights from \underline{\textbf{T}}ext, temporal information, and \underline{\textbf{S}}patio-\underline{\textbf{T}}emporal data from mobility sensors (TST). Such integration significantly improves prediction performance, and can serve as a unified framework for addressing tasks within dynamic urban environments where external events frequently reshape mobility patterns. The architecture is designed to be lightweight and efficient, supporting the minute-level responsiveness required for real-time applications.
Our contribution can be summarized as follows:
\begin{itemize}
    \item We design a multi-agent framework for automated extraction and reasoning of event-related textual context from online sources for urban mobility analysis.
    \item We create a unique context-enriched dataset for mobility forecasting, aligning event-related textual narratives with fine-grained spatiotemporal mobility data.
    \item We propose a progressive fusion architecture that dynamically weights multimodal inputs. Our findings demonstrate the significant benefits of integrating textual information for event-related mobility prediction.
\end{itemize}

\section{Preliminaries}
Let $N$ be the total number of sensors in a monitored urban area. The sensor network is represented by a graph $\mathcal{G} = (\mathcal{V}, \mathcal{E}, \mathbf{A})$, where $\mathcal{V}= \{v_1, \dots, v_N\}$ is the set of $N$ sensors, $\mathcal{E}$ is the set of edges representing connectivity, and $\mathbf{A} \in \mathbb{R}^{N \times N}$ is an adjacency matrix encoding spatial relationships (e.g., based on pairwise geodesic distances) among sensors.

\paragraph{Traditional Spatiotemporal Mobility Prediction}
At any given time step $t$, the historical mobility flow signals from all $N$ sensors over the past $T$ time slices are denoted by a tensor $\mathbf{X}_{[t-T+1:t]} \in \mathbb{R}^{N \times T}=\left(\mathbf{x}_{t-T+1}, \mathbf{x}_{t-T+2}, \dots, \mathbf{x}_t\right)^\top$. Each $\mathbf{x}_i \in \mathbb{R}^{N}$ captures the flow data across all $N$ sensors at time slice $i$.

Traditional spatiotemporal mobility prediction is to learn a mapping function $f$ that, given the historical flow data $\mathbf{X}_{[t-T+1:t]}$ and the sensor network graph $\mathcal{G}$, predicts the mobility flows for the subsequent $T'$ time slices:
\begin{equation}
f: (\mathbf{X}_{[t-T+1:t]}, \mathcal{G}) \mapsto \hat{\mathbf{X}}_{[t+1:t+T']} \nonumber
\end{equation}
where $\hat{\mathbf{X}}_{[t+1:t+T']} \in \mathbb{R}^{N \times T'}$ is the sequence of predicted future flow signals.

\paragraph{Event-driven Spatiotemporal Mobility Prediction} We consider an event occurring at a specific venue, where $V_M\subset V$ is a subset of $M$ sensors ($M\le N$) identified as being affected by the event. At any given time step $t$ within the event's impact window, the historical mobility flow signals for these $M$ affected sensors over the past $T$ time slices are denoted by $\mathbf{X}^{(M)}_{[t-T+1:t]}\in\mathbb{R}^{M\times T}$. Each sensor is associated with a set of spatial relationship features relative to the event venue, forming a matrix $\mathbf{D}\in \mathbb{R}^{M \times K_D}$, where~$K_D$~is the number of distinct spatial features (e.g., distance to venue, orientation relative to venue).
Additionally, event-specific textual information $\mathcal{T}_{event}$ is established and finalized prior to the event day.

Extending from traditional spatiotemporal mobility prediction, the mapping function needs to condition further on venue-related features $\mathbf{D}$ and event information $\mathcal{T}_{event}$. Event-driven spatiotemporal mobility prediction is to learn a mapping function $g$ that maps historical affected sensor flows $\mathbf{X}^{(M)}_{[t-T+1:t]}$, venue-related features $\mathbf{D}$, event information $\mathcal{T}_{event}$, and the broader network $\mathcal{G}$ to the predicted $T'$-slice future flows $\hat{\mathbf{X}}^{(M)}_{[t+1:t+T']}\in \mathbb{R}^{M \times T'}$:
\begin{equation}
g: (\mathbf{X}^{(M)}_{[t-T+1:t]}, \mathbf{D}, \mathcal{T}_{event}, \mathcal{G}) \mapsto \hat{\mathbf{X}}^{(M)}_{[t+1:t+T']} \nonumber
\end{equation}
This formulation models practical scenarios that aim to enhance real-time mobility predictions during the event by leveraging comprehensive event-specific information gathered beforehand.

\section{Methodology}
 The overall architecture of SeMob, detailing the workflow and the specific roles of agents, is illustrated in Figure~\ref{fig:multi-agent}. 

\subsection{Multi-agent Framework}
\label{section:agent}
 Agents extract, filter, and reason about event texts relevant to spatiotemporal mobility through the following specialized modules:
\paragraph{Information Retrieval Module} This module gathers multi-dimensional event information through two specialized agents: \textbf{E}vent \textbf{I}nfo Extractor (\textbf{EI}) and \textbf{T}weet \textbf{A}nalyzer (\textbf{TA}). \textbf{EI} gathers basic event information from the official venue calendar database, such as time and location. This agent summarizes the event content and conducts a preliminary analysis of the event scale and target audience. \textbf{TA} uses the basic information provided by \textbf{EI} to construct retrieval keywords for searching relevant tweets from the month preceding the event. After filtering tweets for relevance, \textbf{TA} extracts detailed event information (such as opening ceremonies) and gauges public interest in the event.
\paragraph{Mobility Reasoning Module} Events affect mobility on various spatiotemporal scales, requiring careful filtering of relevant information. This module employs a \textbf{M}obility \textbf{A}nalyzer (\textbf{MA}) agent to reason texts with high spatiotemporal relevance to potential mobility impacts. For any given day of events under analysis, the \textbf{MA} considers both current-day and recent proximate events. It applies a predefined information-mobility correlation logic to identify textual evidence of spatiotemporal traffic patterns around venues during event periods. The output comprises texts detailing these impact patterns, associated core event information, and relevant public reactions.
\paragraph{Evaluation Module} Developing a perfect information screening logic a priori is challenging even for domain experts. The \textbf{Evaluator} in this module analyzes instances where mobility predictions surrounding a venue exhibit significant errors for specific time periods or locations. By examining these cases, the agent identifies potentially overlooked information or misjudged impact factors. This analysis is fed back into the system, updating the screening logic for similar scenarios. For detailed prompts for these agents, see Appendix~\ref{APP:prompt}.

\subsection{Multimodal Fusion}
The TST module achieves event-driven spatiotemporal mobility prediction through a two-stage progressive fusion of multimodal signals, as shown in Figure~\ref{fig:multi}. Further details are provided in Appendix~\ref{App:Model details}.

\begin{figure}[tbp]
  \centering 
  \includegraphics[width=\linewidth]{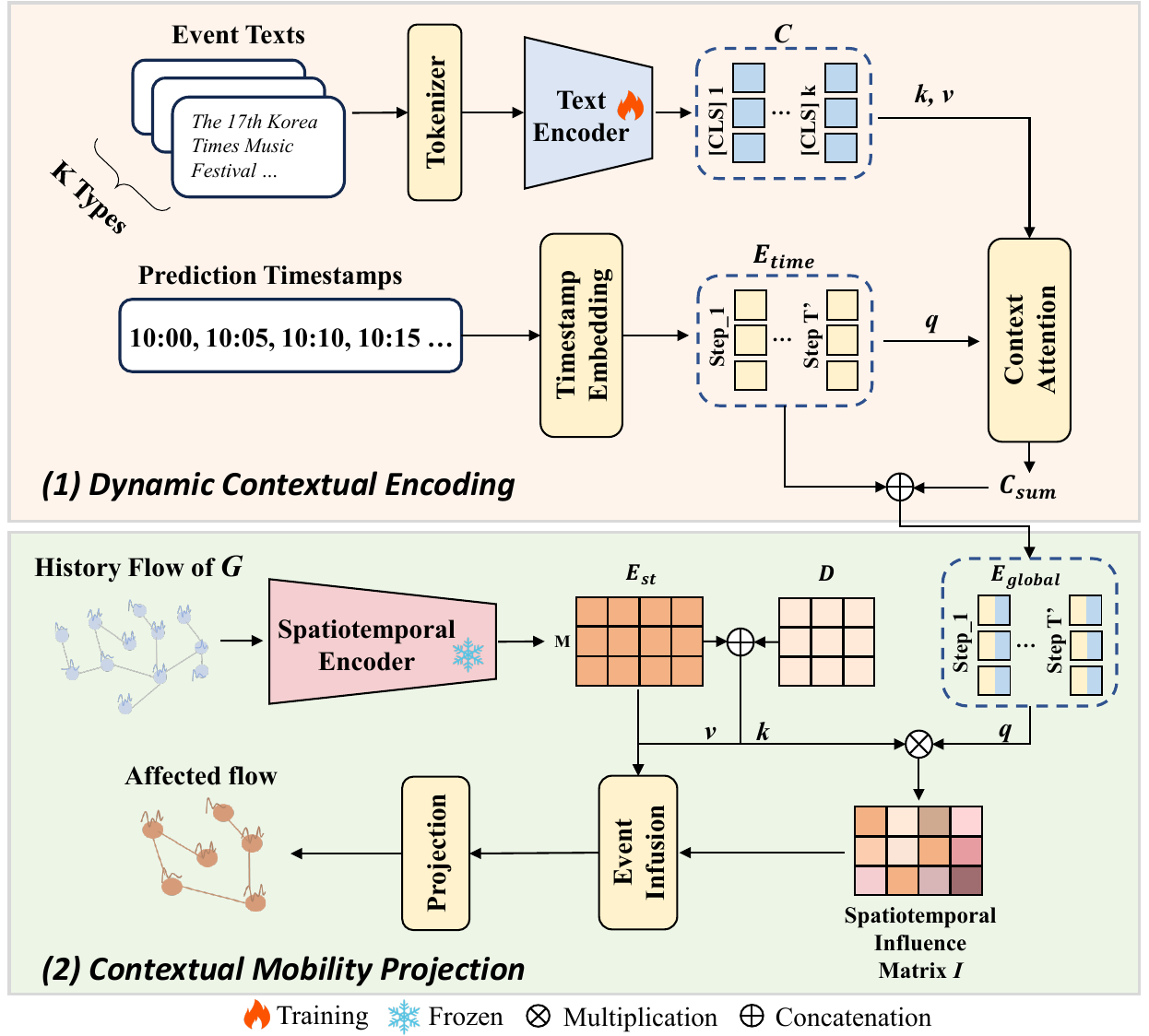}
  \caption{TST architecture. We first (1) synthesizes dynamic event signatures by fusing textual embeddings with evolving temporal contexts and (2) integrates spatiotemporal data with these signatures for fine-grained, context-aware mobility predictions.}

  \label{fig:multi}
\end{figure}

\paragraph{Dynamic Contextual Encoding.} 
The first stage distills a sequence of dynamic event signatures, $\bm{E}_{global} \in \mathbb{R}^{T' \times d_g}$, by contextualizing an initially static textual essence with temporal information for $T'$ prediction steps. We begin by sourcing $K$ distinct categories of event-related text, which are identified by the multi-agent workflow in section~\ref{section:agent}. These texts are represented as a content embedding matrix $\bm{C} \in \mathbb{R}^{K \times d_h}$, where each row corresponds to the [CLS] token from a tunable RoBERTa encoder~\citep{liu2019roberta}. For each future prediction step $t \in [1, T']$, an learnable embedding of its corresponding timestamp, $\bm{e}_{time}^{t} \in \mathbb{R}^{d_t}$, provides the specific temporal context. To integrate textual information under this temporal setting, we employ $f_{attn}$: a context-driven attention mechanism. $f_{attn}$ uses $\bm{e}_{time}^{t}$ as a query to process the static category embeddings $\bm{C}$ (which serve as keys and values), producing a temporally-focused textual summary $\bm{c}_{sum}^{t}$. This summary reflects how the diverse textual facets combine under the temporal lens of step $t$. The event signature for this step, $\bm{E}_{global}^{t}$, is then formed by concatenating $\bm{c}_{sum}^{t}$ with $\bm{e}_{time}^{t}$, followed by an FFN transformation:
\begin{equation} \label{eq:c_sum}
\bm{c}_{sum}^{t} = f_{attn}(\bm{e}_{time}^{t}, \bm{C}) 
\end{equation}
\begin{equation} \label{eq:e_global_t_formation}
\bm{E}_{global}^{t} = \bm{W}[\bm{c}_{sum}^{t} \oplus \bm{e}_{time}^{t} ]+ \bm{b}
\end{equation}
where $\oplus$ denotes concatenation. $\bm{W}$ and $\bm{b}$ are the learnable parameters. The collection of $T'$ step-specific event signatures sequence $\bm{E}_{global}$ offers a dynamically evolving, temporally-contextualized representation of the event across the entire prediction horizon.

\paragraph{Contextual Mobility Projection.}
To infuse the event-aware information into the spatiotemporal data, we design a cross-modal integration mechanism. For $M$ affected sensors, their initial spatiotemporal embeddings $\bm{E}_{st}$ (from a pre-trained spatiotemporal encoder on network $\mathcal{G}$) and spatial features $\mathbf{D}$ are concatenated to form per-sensor local spatiotemporal context representations $\bm{S}_{loc}$.  We employ two linear layers, $f_q$ and $f_k$ to transform  $\bm{E}_{global}$ and $\bm{S}_{loc}$ into two compact embeddings: $f_q(\bm{E}_{global})$ and $ f_k(\bm{S}_{loc})$. Spatiotemporal influence weights $\bm{I} \in \mathbb{R}^{M \times T' }$ are then computed by matrix multiplication followed by softmax:
\begin{equation} \label{eq:influence_weights_A_MT}
\bm{I^T} = \text{Softmax}\left( \frac{f_q(\bm{E}_{global}) \cdot (f_k(\bm{S}_{loc})^T}{\sqrt{d_k}} \right)
\end{equation}
where $d_k$ is the dimensionality of keys and queries. For each sensor $i$ and future step $t$, the event-infused representation $\bm{Z}_{i,t}$ combines its initial spatiotemporal embedding $\bm{E_{st}^i}$ with $\bm{V}_i$ scaled by the influence weight $\bm{I}_{i,t}$. The value embedding $\bm{V}_i$ is given by $f_{value}(\bm{E_{st}^i})$, where $f_{value}$ is a learnable projection preserving the input feature dimension.
\begin{equation} \label{eq:event_infused_representation_Z_it_activated}
\bm{Z}_{i,t} = E_{st}^i + \bm{I}_{i,t} \cdot \sigma(\bm{V}_i)
\end{equation}
where $\sigma$ is an activation function.The event-infused representations for $M$ sensors are subsequently processed by a one-layer FFN to project into the final multi-step mobility flow predictions $\hat{\mathbf{X}}\in \mathbb{R}^{M \times T'}$. 
Through cross-modal integration, we extend the predictive capability to dynamically incorporate semantics for fine-grained, context-sensitive spatiotemporal forecasting under information-driven dynamics.

\section{Experiments}
\subsection{Experiment Setup}
\paragraph{Data Preparation}
The multi-agent workflow framework retrieves information from venue databases and analyzes tweets posted within 30 days prior to each event\footnote{\url{x.com/search-advanced}}. Specific venue information sources include official websites (e.g., cryptoarena.com, rosebowlstadium.com) and local event aggregators (e.g., dolosangeles.com, sfgate.com/events). The prompts used to guide the agents for tweet filtering and text synthesis are detailed in Appendix~\ref{APP:prompt_event} and~\ref{APP:prompt_tweet}. 911 events of various types occurring over a full year are collected for analysis. To capture the corresponding mobility dynamics, we then collect traffic flow data from the Caltrans Performance Measurement System (PeMS\footnote{\url{pems.dot.ca.gov}}) at a 5-minute temporal resolution. 

We selected the major Californian event hubs that simultaneously offered high-quality PeMS data, accessible event records, and a direct location along PeMS-monitored road networks. We gather data from sensors located within 2km, 3km, 4km, and 5km radii of each venue. These specific boundaries were established to create challenging, proximate zones for model validation, rather than to define an event's absolute area of influence. The inner radius of 2km guarantees sensor coverage for all venues, whereas the 5km outer boundary is empirically set based on our analysis in Figure~\ref{fig:spatiotemporal}. Furthermore, to examine temporal impacts, the sensor data is analyzed within time windows spanning from two, three, or four hours before each event to a corresponding duration afterward. The dataset is partitioned by chronological order and event type. More details are provided in Appendix~\ref{App:Dataset}.

\paragraph{Baselines and Evaluation}
We evaluate our model against nine leading traffic forecasting baselines on large-scale road networks~\citep{liu2023largest,wang2024stone}: (1) temporal-only methods: LSTM~\citep{fu2016lstm} and PatchTST~\citep{Yuqietal2023PatchTST}; (2) GNN-RNN based models: DCRNN~\citep{li2018dcrnn} and AGCRN~\citep{bai2020agcrn}; (3) GNN-TCN based models: STGCN~\citep{yu2018stgcn} and GWNET~\citep{wu2019gwnet}; (4) attention-based method: ASTGCN~\citep{guo2019astgcn}; (5) ordinary differential equation based model: STGODE~\citep{fang2021spatial}; and (6) dynamic graph based approach: DSTAGNN~\citep{lan2022dstagnn}. These spatiotemporal models are trained on the entire road network graph within the broader region where the venue is located, while testing performance specifically on affected sensors during event windows. The evaluation metrics are mean square error (MSE) and mean absolute error (MAE). We utilize 12 historical steps to predict 12 future steps (predicting the next hour based on the previous hour), consistent with established prediction benchmarks. More details can be found in Appendix~\ref{App:Experimental Setting}.

\begin{table*}[ht]
    \centering
    \small
    \setlength{\tabcolsep}{4pt}
    \renewcommand{\arraystretch}{0.9}
    \begin{tabular}{lcccccccccccc}
        \toprule
        \multirow{3}[4]{*}{\textbf{Methods}} & \multicolumn{6}{c}{\textbf{By Time}} & \multicolumn{6}{c}{\textbf{By Type}} \\
        \cmidrule(lr){2-7} \cmidrule(lr){8-13} 
        & \multicolumn{2}{c}{\textbf{2h}} & \multicolumn{2}{c}{\textbf{3h}} & \multicolumn{2}{c}{\textbf{4h}} & \multicolumn{2}{c}{\textbf{2h}} & \multicolumn{2}{c}{\textbf{3h}} & \multicolumn{2}{c}{\textbf{4h}} \\
        \cmidrule(lr){2-3} \cmidrule(lr){4-5} \cmidrule(lr){6-7} \cmidrule(lr){8-9} \cmidrule(lr){10-11} \cmidrule(lr){12-13}
        & \textbf{MAE} & \textbf{RMSE} & \textbf{MAE} & \textbf{RMSE} & \textbf{MAE} & \textbf{RMSE} & \textbf{MAE} & \textbf{RMSE} & \textbf{MAE} & \textbf{RMSE} & \textbf{MAE} & \textbf{RMSE}\\
        \midrule
        LSTM & 30.05 & 41.44 & 27.65 & 40.23 & 27.12 & 39.71 & 30.33 & 42.60 & \underline{27.73} & 41.10 & 27.16 & 40.45 \\
        PatchTST & 31.22 & 41.90 & 27.86 & 41.00 & 27.25 & 40.74 & 31.55 & 43.86 & 27.95 & 42.09 & 27.29 & 41.57 \\
        DCRNN & 30.06 & 41.23 & 29.06 & 41.01 & 28.55 & 40.88 & 30.76 & 42.78 & 29.43 & 42.12 & 28.76 & 41.77 \\
        AGCRN & 32.43 & 47.06 & 30.62 & 49.23 & 30.33 & 52.62 & 31.44 & 45.51 & 29.19 & 46.37 & 28.82 & 47.71 \\
        STGCN & 35.40 & 52.69 & 33.12 & 52.36 & 32.40 & 52.21 & 34.79 & 51.77 & 31.92 & 49.79 & 31.02 & 48.92 \\
        GWNET & \underline{29.03} & \underline{39.92} & \underline{27.01} & \underline{38.61} & \underline{25.64} & \underline{37.35} & \underline{29.13} & \underline{40.83} & 27.77 & \underline{39.33} & \underline{25.69} & \underline{37.88} \\
        ASTGCN & 42.16 & 60.86 & 40.37 & 59.20 & 39.36 & 58.33 & 40.45 & 60.20 & 37.58 & 58.11 & 35.97 & 57.02 \\
        STGODE & 35.21 & 51.95 & 32.96 & 50.37 & 31.88 & 49.58 & 34.64 & 51.62 & 31.93 & 49.56 & 30.63 & 48.53 \\
        DSTAGNN & 30.77 & 42.01 & 29.71 & 41.85 & 29.27 & 41.78 & 31.18 & 43.64 & 29.60 & 42.87 & 28.95 & 42.56 \\
        \textbf{TST} & \textbf{25.13} & \textbf{36.71} & \textbf{24.75} & \textbf{36.34} & \textbf{23.82} & \textbf{35.78} & \textbf{25.08} & \textbf{36.29} & \textbf{24.59} & \textbf{36.07} & \textbf{23.77} & \textbf{35.56} \\
        \bottomrule
    \end{tabular}
    \caption{Performance comparison of baselines using sensor data within 3 km of event locations. The \textbf{best} and the \underline{second-best} results are indicated by bold and underlined text respectively. Datasets are partitioned either chronologically ('By Time') or by event type ('By Type').}
    \label{tab:main}
\end{table*}

\subsection{Main Results}
\label{comparision}
The performance of our method compared to spatiotemporal models is shown in Table~\ref{tab:main}.We can observe from the experimental results that:

\paragraph{Event-specific information yields superior performance.}
Our approach significantly outperforms traditional methods that rely solely on time series data and spatial relationships. Incorporated by event context, the model achieves maximal reductions of 13.92\% in MAE and 11.12\% in RMSE relative to the best-performing baseline spatiotemporal model. Evaluation on datasets partitioned by event type reveals slightly greater performance enhancements compared to chronological partitioning. Such partitioning facilitates more balanced learning across event categories, allowing the model to capture category-specific mobility patterns more effectively.

\paragraph{Spatiotemporal sensitivity confirms the importance of event context.} Detailed in Appendix~\ref{App:Experiment results}, conventional models exhibit degraded performance and heightened sensitivity closer to event venues and times. The observed degradation in accuracy indicates that human mobility during events becomes diverging from regular patterns and proving challenging for models to predict accurately. Conversely, our proposed method leverages relevant texts to achieve pronounced superiority precisely within these highly affected spatiotemporal scales.

\subsection{Ablation Study}
We compare our full model against six distinct variants targeting fusion strategies and multi-modal encoders: (1)~\textit{w/}~EF: early fusion via concatenation; (2)~\textit{w/}~LF: late fusion post-independent feature extraction; (3)~\textit{w/o}~TT: replacing dynamic contextual encoding fusion with concatenation; (4)~\textit{w/o}~TS: interaction between the global representation and spatiotemporal data substituted with concatenation; (5)~\textit{w/o}~Finetune: frozen pre-trained text encoders; and (6)~\textit{w/o}~STE: processing spatiotemporal data without pre-trained encoders. The experiments utilize data within 3km of event locations across 2h, 3h, and 4h windows and the results are shown in Table~\ref{tab:ablation}.
Our progressive fusion strategy demonstrates superiority over early and late fusion, revealing limitations of simple feature concatenation for capturing complex text-spatiotemporal relationships. Performance degradation when removing dynamic contextual encoding and global-sensor interactions confirms the necessity of the multi-level interaction mechanisms for capturing cross-modal dependencies. The most significant drop occurs when utilizing a frozen decoder, which indicates adapting encoders is crucial for extracting spatiotemporally relevant semantic features rather than relying on generic representations. Excluding pre-trained spatiotemporal encoders also impairs predictive performance, affirming their value in capturing latent patterns. Interestingly, the performance decline is less severe than other ablations, suggesting the primacy of fused and adapted textual information for event-related prediction.
\begin{table}[ht]
    \centering
    \small
    \setlength{\tabcolsep}{2pt}
    \renewcommand{\arraystretch}{0.9}
    \begin{tabular}{lcccccc}
        \toprule
        \multirow{2}[2]{*}{\textbf{Methods}} & \multicolumn{2}{c}{\textbf{2h}} & \multicolumn{2}{c}{\textbf{3h}} & \multicolumn{2}{c}{\textbf{4h}} \\
        \cmidrule(lr){2-3} \cmidrule(lr){4-5} \cmidrule(lr){6-7}
        & \textbf{MAE} & \textbf{RMSE} & \textbf{MAE} & \textbf{RMSE} & \textbf{MAE} & \textbf{RMSE} \\
        \midrule
        \textbf{TST}          & \textbf{25.08} & \textbf{36.29} & \textbf{24.59} & \textbf{36.07} & \textbf{23.77} & \textbf{35.56} \\
        \textit{w/}~EF        & 26.87 & 38.86 & 25.42 & 37.63 & 24.29 & 36.47 \\
        \textit{w/}~LF        & 26.98 & 39.12 & 25.46 & 37.82 & 24.32 & 36.58 \\
       \textit{w/o}~TT       & 25.97 & 38.03 & 25.09 & 37.00 & 24.11 & 36.19 \\
        \textit{w/o}~TS       & 26.19 & 38.69 & 25.29 & 37.74 & 24.24 & 36.57 \\
        \midrule
       \textit{w/o}~Finetune & 27.04 & 38.95 & 25.50 & 37.70 & 24.34 & 36.51 \\
        \textit{w/o}~STE      & 25.86 & 37.90 & 24.89 & 36.94 & 23.96 & 36.07 \\
        \bottomrule
    \end{tabular}
    \caption{Ablation experiments on fusion strategies and multi-modal encoders.}
    \label{tab:ablation}
\end{table}
\section{Discussion}
\begin{figure}[htbp]
  \centering 
  \includegraphics[width=\linewidth]{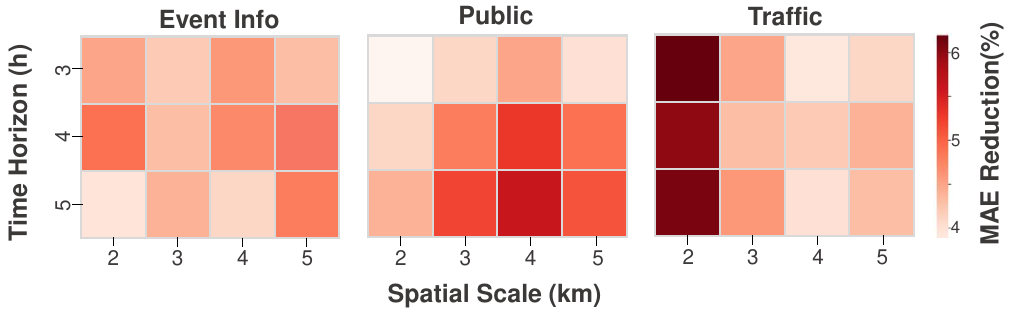}
  \caption{Comparative visualization of MAE reduction derived from basic event information, public reaction, and inferred traffic conditions versus spatiotemporal information alone.}
  \label{fig:text}
\end{figure}

In this section, we discuss the following research questions (RQ) of the proposed pipeline:
\begin{itemize}
    \item \textbf{RQ1}: To what extent does the contextual information enhance mobility prediction?
    \item \textbf{RQ2}: What are the contributions of the roles and stages within the multi-agent framework to the forecast results?
    \item \textbf{RQ3}: Can the system still effectively leverage the textual context when the scale of the training events becomes smaller?
    \item \textbf{RQ4}:  Why can our method enable the effective integration of event-related textual context with spatiotemporal data?

\end{itemize}
\paragraph{Response to RQ1: Different categories of textual information provide varied yet consistently beneficial contributions to prediction across spatiotemporal scales.} We investigate the contributions of different textual features and present visualizations in Figure~\ref{fig:text}. Although all textual categories enhance the predictive accuracy of the spatiotemporal model, their specific impacts diverge. Basic event information offers relatively stable performance improvements across all spatiotemporal granularities. Public reaction provides more specific advantages, yielding significant benefits for forecasts closer to event venues and over extended event windows. The inferred traffic conditions primarily improve the prediction accuracy for sensors near event venues. The localized efficacy of such an inference may stem from the agent's limited capability to analyze near-venue impacts. According to the findings, synthesizing information from different textual sources is advantageous for capturing complex, non-linear mobility patterns.

\paragraph{Response to RQ2: Collaboration among specialized agents generates high-quality textual context for model training.}
To validate the contributions of inter-agent collaboration, we compare key workflow configurations.
Results in Table~\ref{tab:agent} reveal that contextual information from source-specific agents provides significantly greater predictive utility when subsequently refined and analyzed by a \textbf{M}obility  \textbf{A}nalyzer. Omitting this analytical synthesis may result in the direct use of less processed information, which correlates with suboptimal downstream performance. Furthermore, the MA's strategy of considering recent event dynamics is important for generating effective training context. Disregarding such temporal information impairs the final prediction accuracy of models trained thereon. Crucially, results also demonstrate that the Evaluator can refine the MA’s logic for text filtering and relevance assessment. The improvement in prediction accuracy after evaluation iterations reflects the dynamic optimization, underscoring the critical role of the Evaluator in maintaining and enhancing the quality of the generated textual context. Collectively, these observations highlight that the multi-agent framework's efficacy stems from its multi-stage collaborative process.
\begin{table}[ht]
    \centering
    \small
    \setlength{\tabcolsep}{2pt}
    \renewcommand{\arraystretch}{0.9}
    \begin{tabular}{lcccccc}
        \toprule
        \multirow{2}[2]{*}{\textbf{Methods}} & \multicolumn{2}{c}{\textbf{2h}} & \multicolumn{2}{c}{\textbf{3h}} & \multicolumn{2}{c}{\textbf{4h}} \\
        \cmidrule(lr){2-3} \cmidrule(lr){4-5} \cmidrule(lr){6-7}
        & \textbf{MAE} & \textbf{RMSE} & \textbf{MAE} & \textbf{RMSE} & \textbf{MAE} & \textbf{RMSE} \\
        \midrule
        \textbf{TST} & \textbf{25.08} & \textbf{36.29} & \textbf{24.59} & \textbf{36.07} & \textbf{23.77} & \textbf{35.56} \\
        w/o MA & 26.49 & 37.97 & 25.63 & 37.33 & 24.46 & 36.79 \\
        w/o RE & 26.15 & 38.23 & 25.41 & 37.68 & 24.43 & 37.02 \\
        w/o Eval & 26.03 & 37.86 & 25.34 & 37.32 & 24.05 & 36.03 \\
        \bottomrule
    \end{tabular}
    \caption{Performance comparison of different multi-agent workflow configurations. "w/o MA" indicates source information not processed through analysis of the Mobility Analyzer. "w/o RE" represents Mobility Analyzer variants operating without considering recent events. "w/o Eval" represents systems without evaluation agent refinement. Experiments utilize data within 3km of event locations across 2h, 3h, and 4h windows.}
    \label{tab:agent}
\end{table}

\begin{figure}[tbp]
  \centering 
  \includegraphics[width=\linewidth]{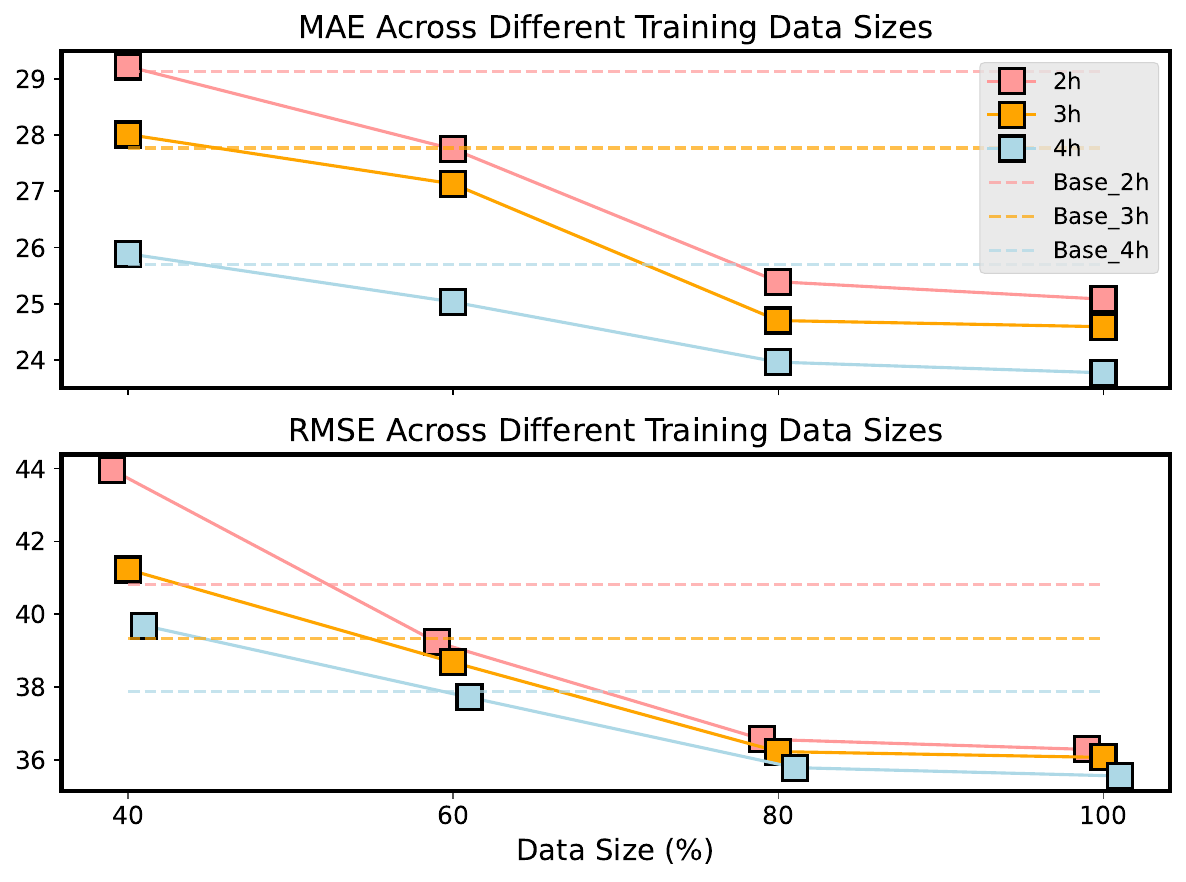}
  \caption{Performance of different time windows across training data sizes. ‘Base’ indicates the performance of GWNET without event information.}
  \label{fig:data}
\end{figure}

\paragraph{Response to RQ3: Yes. Even with limited data, our model can enhance the mobility prediction.} We compare the performance of models trained with more limited quantities of training events, and present the results in Figure~\ref{fig:data}. Although models generally achieve better performance with a larger training dataset, even with only 60\% of the training data, our model can still outperform the base model that solely utilize spatiotemporal data. 

\paragraph{Response to RQ4: Multimodal fusion-guided fine-tuning enables language models to extract spatiotemporal dynamic-related features from raw text.} To quantitatively evaluate this capability, we examine differences in [CLS] token attention patterns between pre-trained and fine-tuned models on test set samples. Specifically, we measure the proportion of pre-defined spatiotemporally relevant words found within the top-10 tokens most attended by each model's [CLS] token. The results presented in Table~\ref{tab:attention} show that fusion-guided fine-tuning redirects model attention towards granular cues indicative of spatiotemporal variations. 
\begin{table}[htbp]
    \centering
    \small 
    \setlength{\tabcolsep}{2pt} 
    \begin{tabular}{lcccc}
    \toprule
          \multirow{2}[2]{*}{\textbf{Source}}& \multicolumn{2}{c}{Pretrained} & \multicolumn{2}{c}{Fine-tuned} \\
\cmidrule(lr){2-3} \cmidrule(lr){4-5} 
 & Spatial  & Temporal  & Spatial & Temporal  \\
\midrule
Event\_info & 36.87        & 32.96         & \textbf{79.32}        & \textbf{75.42}         \\
Public      & 16.76        & 27.37         & \textbf{46.37}        & \textbf{54.75}         \\
\bottomrule
\end{tabular}
 \caption{Proportion (\%) of pre-defined spatial and temporal words within the top-10 `[CLS]`-attended tokens from pre-trained and fusion-guided fine-tuned models.}
\label{tab:attention} 
\end{table}

\begin{figure}[tpb]
  \centering 
  \includegraphics[width=\linewidth]{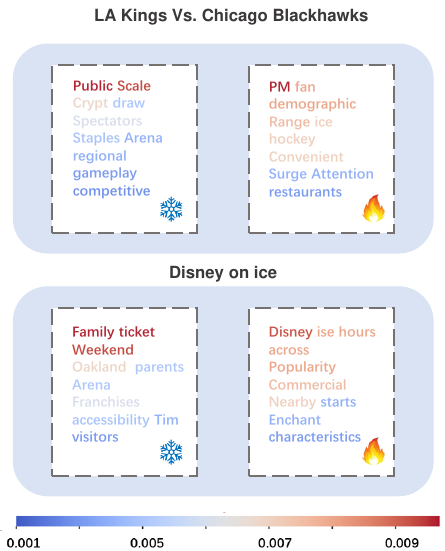}
  \caption{Examples of [CLS] token attention. The visualization contrasts the top 10 words most attended by the [CLS] token from pre-trained (left) and fine-tuned (right) models processing the same event information.}
  \label{fig:sample}
\end{figure}

Figure~\ref{fig:sample} offers qualitative support, providing two illustrative examples of [CLS] token attention visualizations. Beyond the general redirection evident from Table~\ref{tab:attention}, examples in Figure~\ref{fig:sample} further reveal the fine-tuned model's improved acuity in identifying distinctive event characteristics (e.g., categorizing sports events as "ice hockey" or entertainment as "Disney" with thematic terms like "enchant"). Such features inherently govern unique spatiotemporal patterns. The attention shift after fine-tuning provides more targeted and impactful inputs for spatiotemporal mobility predictions.

\section{Related Work}
\paragraph{Human Mobility Spatiotemporal Prediction}
Dominant approaches in this field include time-series forecasting techniques~\citep{fu2016lstm, zhou2021informer, Yuqietal2023PatchTST} and hybrid architectures combining Graph Convolutional Networks with sequence models~\citep{li2018dcrnn, wu2019gwnet, bai2020agcrn}. Attention mechanisms and dynamic graphs represent further refinements for modeling evolving spatial interactions~\citep{guo2019astgcn, lan2022dstagnn, gravina2024deep}.
These methods utilize structured data, consequently overlooking information in unstructured text (e.g., social media, event sites) critical for capturing event-driven mobility shifts. 

While some studies have attempted to incorporate textual data, they typically rely on manual, task-specific feature engineering that discretizes text into categorical variables~\citep{tu2023multi, han2024event}. Such an approach not only suffers from significant semantic loss but also lacks the scalability required for diverse real-world scenarios with rich event descriptions. Furthermore, this line of work is often applied to constrained problem settings, such as single-station passenger flow or localized event impact, failing to capture the network-wide mobility shifts that are our focus.
\paragraph{Multimodal Fusion}
Cross-modal learning with textual data has successfully addressed numerous real-world audiovisual understanding tasks~\citep{huang2024vtimellm,cai2024av}. Building on these advances, recent research has extended text-based approaches to time-series data interpretation through LLMs~\citep{gruver2023large,liu2024autotimes,hu2025context}. However, these methods rely on extensive aligned datasets for pretraining, such as text-image pairs or time-series-text description pairs. However, for event-driven spatiotemporal data, textual information remains scarce and poorly aligned with fine-grained spatiotemporal signals. Furthermore, the high-dimensional nature of spatiotemporal data creates fundamental alignment challenges with sequential textual representations~\citep{jin2023large}. Our work therefore advances multi-modal fusion techniques to specifically address these limitations. Multimodal fusion encompasses diverse strategies including input-level, representation-level, and prediction-level~\citep{xu2023multimodal}. Selecting an appropriate fusion strategy is critical to meet the low-latency requirements of operational mobility prediction.
\paragraph{Multi-Agent Framework}The advancement of LLMs powered agents has fostered significant advances in complex task resolution through human-like capabilities, including retrieval augmentation~\citep{asai2023self}, role-playing~\citep{park2023generative} and communication~\citep{park2024predict}. By coordinating diverse agent capabilities and roles within multi-agent frameworks, systems can address challenging problems through structured collaboration~\citep{chan2023chateval,hong2024metagpt}. This paradigm has demonstrated remarkable efficacy across domains such as healthcare~\citep{qiu2024llm,wang2025colacare} and finance~\citep{yu2024fincon}. Urban mobility analysis represents a collaborative task that requires information retrieval and filtering. Our designed multi-agent workflow facilitates this teamwork approach and can provide effective descriptive data for prediction tasks.
\section{Conclusion}
In this paper, we first identify the critical challenge of human mobility forecasting under event-driven dynamics. We propose SeMob, an LLM-powered semantic synthesis framework with a multi-agent system for extracting mobility-relevant event information and a progressive architecture for fusing textual and spatiotemporal data. Experimental results demonstrate that SeMob achieves superior forecasting accuracy, particularly  within highly event-affected spatiotemporal scales.

\section*{Limitations}
While this study demonstrates a novel approach to the integration of textual data for the prediction of spatiotemporal mobility, certain limitations should be acknowledged. The framework's effectiveness is most evident for large planned events with abundant textual data; its application to smaller or less-documented events with sparse text requires further exploration. Specifically, our methodology is designed for the paradigm of "prediction and planning" ahead of large-scale scheduled events. Consequently, it is not directly applicable to ad-hoc or emergency scenarios, which operate under a different paradigm of real-time "interruption and intervention" and typically lack the advance textual information our model relies on. Extending our semantic fusion paradigm to such sudden events remains a promising direction for future work. Furthermore, this study does not include a comparative evaluation of different LLM for the agent components.
\section*{Ethics  Considerations}
\paragraph{Potential Risks} Although SeMob shows promising potential to streamline event information screening and extract vital insights, we must recognize its inherent constraints. The LLM agents in the pipeline may struggle when confronted with unconventional mobility-related information, sometimes resulting in partial analyses or outputs that require expert interpretation to avoid misunderstanding. Therefore, this system is not intended to replace the expertise of seasoned traffic management professionals or to make autonomous operational decisions. It serves as a supplementary tool to provide data-driven perspectives that assist decision-making processes of traffic managers.
\paragraph{Data Ethics and Privacy Compliance } All data utilized in this work are processed with rigorous attention to privacy and ethical standards. Mobility datasets from PEMS are inherently anonymous. Social media data, also sourced from public platforms, undergoes systematic anonymization of user identifiers to protect personal privacy. Our primary objective is to derive insights relevant to urban mobility, rather than analyze individual data points or behaviors; consequently, the dataset is curated to exclude sensitive or harmful content. Both mobility and social media data are drawn from publicly accessible sources.

\section*{Acknowledgment}
The research was supported by National Natural Science Foundation of China (42171452).
\bibliography{main}
\clearpage
\appendix

\section{Dataset Details}
\label{App:Dataset}
The detailed information for our event-spatiotemporal dataset is shown in Table~\ref{tab:venue_sensors} and~\ref{tab:event_category_distribution}. 
The dataset is partitioned chronologically and by event type. An 8:2 ratio is used for the training and testing sets. We strive to maintain this ratio split within each event type. Crucially, events from the same venue on the same day are kept together in a single set (training or test) to ensure they are not separated. This dataset predominantly features English-language content, which is representative of the primary language used in official event communications and public social media discussions within the geographical scope of our study.
\begin{table}[ht]
    \centering
    \small 
    \setlength{\tabcolsep}{2pt} 
    \begin{tabular}{l>{\centering\arraybackslash}m{1.5cm}ccccc}
        \toprule
        \makecell{\textbf{Venue}} & \textbf{Events} & \textbf{ME} & \textbf{S2} & \textbf{S3} & \textbf{S4} & \textbf{S5} \\
        \midrule
        \makecell{Crypto.com Arena \\ \& LA Convention Center} & 210 & 2 & 22 & 40 & 63 & 87 \\
        \makecell{Rose Bowl Stadium} & 77 & 2 & 13 & 23 & 29 & 37 \\
        \makecell{Hollywood Bowl} & 86 & 1 & 12 & 23 & 31 & 44 \\
        \makecell{The Greek Theatre} & 73 & 2 & 5 & 11 & 36 & 46 \\
        \makecell{Dodger Stadium} & 87 & 2 & 21 & 42 & 64 & 82 \\
        \makecell{Honda Center} & 110 & 4 & 16 & 43 & 76 & 131 \\
        \makecell{Levi's Stadium} & 19 & 2 & 9 & 23 & 35 & 43 \\
       \makecell{Shoreline Amphitheatre} & 37 & 1 & 7 & 13 & 21 & 30 \\
        \makecell{Oakland Arena} & 48 & 1 & 6 & 14 & 18 & 37 \\
        \makecell{SAP Center} & 164 & 3 & 23 & 48 & 82 & 131 \\
        \bottomrule
    \end{tabular}
    \caption{Details of venues and surrounding sensor distributions. ME represents the maximum number of events within a day. S2, S3, S4, and S5 denote the number of sensors within 2 km, 3 km, 4 km, and 5 km radius around each venue, respectively. Levi’s Stadium, Shoreline Amphitheatre, SAP Center, and Oakland Arena are located in the Greater Bay Area, while others are located in the Greater Los Angeles Area. The collected dataset covers the entirety of 2019.}
    \label{tab:venue_sensors}
\end{table}
\begin{table}[ht]
    \centering
    \small
    \begin{tabular}{lc}
        \toprule
        \textbf{Type} & \textbf{Number} \\
        \midrule
        Trade \& Industry   & 56 \\
        Entertainment       & 62 \\
        Celebration         & 97 \\
        Public Service      & 78 \\
        Performing Arts     & 284 \\
        Sports              & 334 \\
        \bottomrule
    \end{tabular}
    \caption{Distribution of event categories. Note that events often belong to multiple categories; this breakdown is intended to visualize the diversity of categories in our collected event set. The distribution, with a higher prevalence of sports and performing arts, reflects the real-world composition of large-scale events in the major Californian metropolitan areas under study.}
    \label{tab:event_category_distribution}
\end{table}

\section{Experimental Setting / Details}
\label{App:Experimental Setting}
\subsection{Baselines} 
\begin{itemize}
\item\textbf{LSTM}~\citep{fu2016lstm}: A classic recurrent neural network designed to capture temporal dependencies through gating mechanisms.
\item\textbf{PatchTST}~\citep{Yuqietal2023PatchTST}: A Transformer-based model that segments time series into patches as input tokens, enabling the capture of local semantic information and long-term dependencies for forecasting. 
\item\textbf{DCRNN}~\citep{li2018dcrnn}: A spatial-temporal model that integrates diffusion graph convolutions with recurrent neural networks to capture spatial dependencies modeled as a diffusion process and temporal dynamics.
\item\textbf{AGCRN}~\citep{bai2020agcrn}: An adaptive graph convolutional recurrent network that learns node-specific patterns and infers inter-dependencies adaptively without a predefined graph structure for traffic forecasting.
\item \textbf{STGCN}~\citep{yu2018stgcn}: A spatial-temporal graph convolutional network that employs graph convolutions to capture spatial structures and 1D convolutions along the time axis to learn temporal features.
\item \textbf{GWNET}~\citep{wu2019gwnet}: A graph WaveNet architecture that combines graph convolutions for spatial feature learning with stacked dilated 1D causal convolutions for temporal dependency modeling.
\item \textbf{ASTGCN}~\citep{guo2019astgcn}: An attention-based spatial-temporal graph convolutional network that utilizes spatial and temporal attention mechanisms alongside graph convolutions to model dynamic spatial-temporal correlations.
\item \textbf{STGODE}~\citep{fang2021spatial}: A model that leverages graph neural networks within an ordinary differential equation framework to capture continuous-time spatial-temporal dynamics.
\item \textbf{DSTAGNN}~\citep{lan2022dstagnn}: A dynamic spatial-temporal aware graph neural network designed to capture evolving spatial dependencies through dynamic graph generation and graph attention mechanisms for improved forecasting accuracy.
\end{itemize}
 Each experiment is independently repeated three times over 100 epochs, with training conducted on traffic data from the year preceding our collected dataset, and the average performance is reported. Model architecture and training configurations follow the recommended settings provided in the official code repositories. 

\subsection{Implementation Detail}
\label{App:Model details}
The multi-agent system incorporates LLM capabilities through Qwen 3\footnote{\url{chat.qwen.ai}}. Each agent within this framework utilizes Chain-of-Thought ~\citep{wei2022chain} and self-reflection~\citep{shinn2023reflexion} techniques. The reflection process is designed to involve three self-iterative cycles of thought to refine the agent's outputs.

In TST module, the timestamp embedding for each prediction step is constructed following the standard practice in spatiotemporal forecasting~\citep{li2018dcrnn, wu2019gwnet,bai2020agcrn}. Specifically, we extract two discrete time features from the timestamp: the day of the week (an integer from 0 to 6) and the time of the day (representing the fraction of the day that has passed). Each of these two features is passed through its own dedicated, trainable embedding layer to be converted into a dense vector. The final timestamp embedding is the concatenation of these two resulting vectors. This method allows the model to effectively learn cyclical temporal patterns from the data.

To fine-tune the text encoder, we adopt the LoRA approach~\citep{hu2022lora} on a RoBERTa model~\citep{liu2019roberta}. We experiment with LoRA ranks from the set \{2, 4, 8\}, keeping the scaling factor $\alpha$ at twice the rank for each, and explore dropout rates of \{0.05, 0.1, 0.2\}. Based on validation performance, we select a rank of 4 (thus $\alpha=8$) and a dropout rate of 0.1 for LoRA fine-tuning. The spatiotemporal encoder leverages pre-trained GWNET embeddings~\citep{wu2019gwnet}, computed from large-scale regional traffic graphs. We employ Smooth L1 Loss~\citep{girshick2015fast} as the objective function. Model training is conducted with 5 epochs using the Adam optimizer~\citep{KingBa15}. The initial learning rate is tuned from the values \{1e-5, 1e-4, 1e-3, 5e-3\}, with 1e-3 being chosen for optimal performance. The batch size is set to 64. All experiments are conducted on an A800 80G GPU. The experiment is repeated three times and the average performance is reported.

Table~\ref{tab:encoder} summarizes the results across different combinations of text and spatiotemporal encoders. More accurate spatiotemporal models yield better representations when used as encoders. Notably, the T5 model~\citep{raffel2020T5}, despite its larger parameter count, underperforms compared to RoBERTa and BERT~\citep{devlin2019bert}. A possible implication is that larger models may introduce unnecessary complexity and irrelevant representations, which could interfere with downstream fusion and prediction tasks. In contrast, since the input text has been distilled by a task-specific agent system to retain only high-quality, relevant information, a lightweight encoder is sufficient to capture the necessary semantics.

\begin{table}[ht]
    \centering
    \small
    \setlength{\tabcolsep}{2pt}
    \renewcommand{\arraystretch}{0.9}
    \begin{tabular}{lcccccc}
        \toprule
        \multirow{2}[2]{*}{\textbf{Methods}} & \multicolumn{2}{c}{\textbf{2h}} & \multicolumn{2}{c}{\textbf{3h}} & \multicolumn{2}{c}{\textbf{4h}} \\
        \cmidrule(lr){2-3} \cmidrule(lr){4-5} \cmidrule(lr){6-7}
        & \textbf{MAE} & \textbf{RMSE} & \textbf{MAE} & \textbf{RMSE} & \textbf{MAE} & \textbf{RMSE} \\
        \midrule
        \textbf{TST}          & \textbf{25.08} & \textbf{36.29} & \textbf{24.59} & \textbf{36.07} & \textbf{23.77} & \textbf{35.56} \\
       \textit{w/}~Bert       & 26.17 & 37.62 & 25.14 & 37.02 & 24.25 & 36.79 \\
        \textit{w/}~T5       & 27.07 & 39.02 & 25.94 & 38.70 & 24.91 & 37.24 \\
        \midrule
       \textit{w/}~DSTAGNN & 25.77 & 37.12 & 24.84 & 36.68 & 23.86 & 35.92 \\
        \textit{w/}~DCRNN      & 25.81 & 37.47 & 24.65 & 36.85 & 23.94 & 36.02 \\
        \bottomrule
    \end{tabular}
    \caption{Ablation experiments on encoder methods. Experiments utilize data within 3km of event locations across 2h, 3h, and 4h windows.}
    \label{tab:encoder}
\end{table}

\section{Comparison of Baselines across Spatiotemporal Scales}
\label{App:Experiment results}
\begin{figure}[tbp]
  \centering 
  \includegraphics[width=\linewidth]{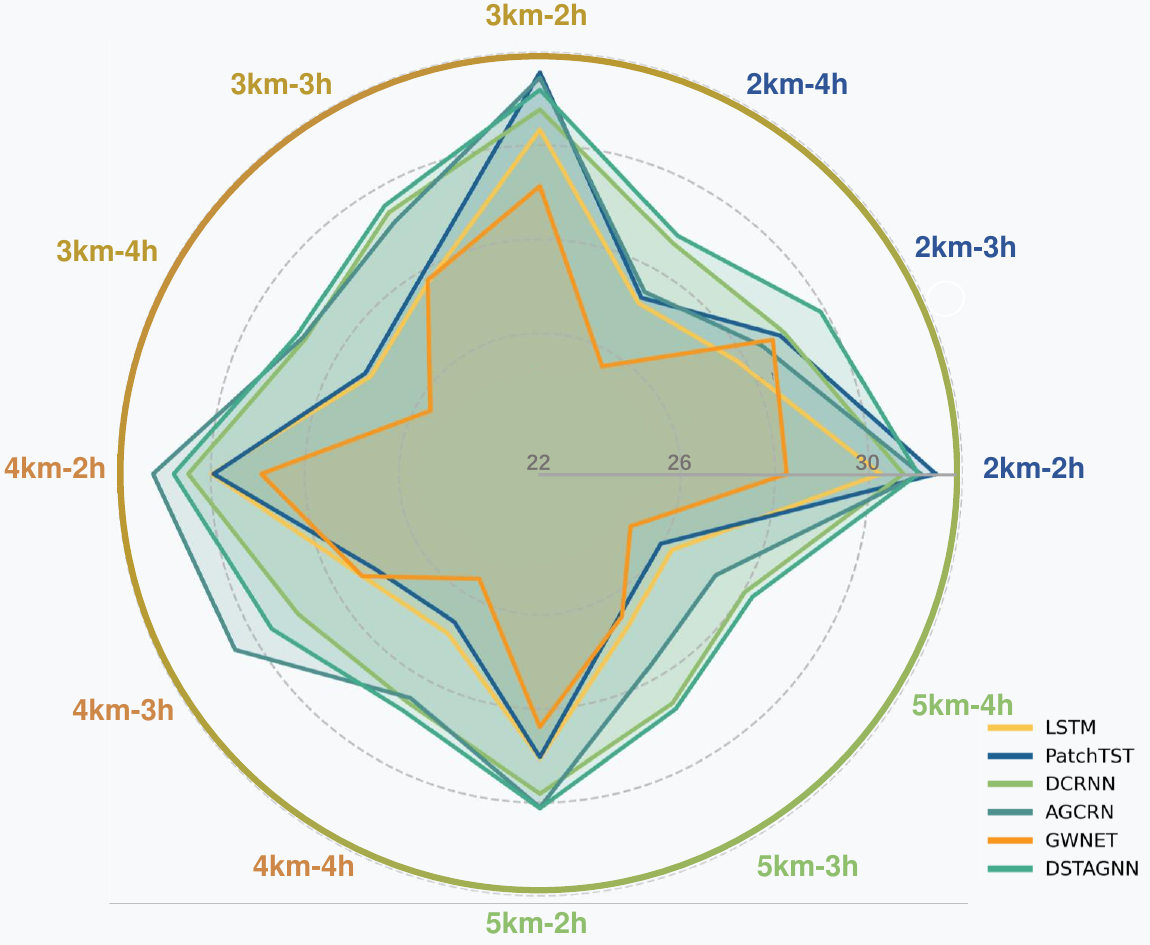}
  \caption{MAE of baselines across spatiotemporal scales.}
  \label{fig:spatiotemporal}
\end{figure}

Detailed performance metrics across diverse spatiotemporal scales are presented in Table~\ref{tab:2km}-~\ref{tab:5km}. Figure~\ref{fig:spatiotemporal} visualizes these results, demonstrating a performance deterioration for most models in the 2km to 3km spatial proximity. In contrast, the rate of performance decrease stabilizes at a 5km radius. Temporal analysis further reveals that a 3-hour event window most significantly influences model accuracy relative to 3-hour and 4-hour intervals.

\section{Performance Metrics for Text Categories}
\label{App: text}
Tables~\ref{tab:mae} and~\ref{tab:rmse} present the MAE and RMSE, respectively, for various text categories evaluated across distances and time intervals.
\begin{table}[htpb]
    \centering
    \small
    \setlength{\tabcolsep}{2pt}
    \renewcommand{\arraystretch}{0.9}
    \begin{tabular}{lccccc}
        \toprule
        \textbf{Category} & \textbf{No\_text} & \textbf{Event\_info} & \textbf{Public} & \textbf{Traffic} & \textbf{TST} \\
        \midrule
        2km\_2h & 28.25 & 26.98 & 27.20 & 26.50 & 25.49 \\
        2km\_3h & 28.72 & 27.31 & 27.54 & 26.99 & 25.32 \\
        2km\_4h & 25.65 & 24.68 & 24.52 & 24.09 & 23.97 \\
        3km\_2h & 29.13 & 27.91 & 27.94 & 27.82 & 25.08 \\
        3km\_3h & 27.77 & 26.58 & 26.44 & 26.57 & 24.59 \\
        3km\_4h & 25.69 & 24.56 & 24.36 & 24.51 & 23.77 \\
        4km\_2h & 28.93 & 27.60 & 27.63 & 27.80 & 24.93 \\
        4km\_3h & 27.36 & 26.07 & 25.91 & 26.21 & 24.22 \\
        4km\_4h & 25.58 & 24.53 & 24.15 & 24.56 & 23.69 \\
        5km\_2h & 28.39 & 27.17 & 27.26 & 27.22 & 24.27 \\
        5km\_3h & 26.50 & 25.15 & 25.20 & 25.33 & 24.07 \\
        5km\_4h & 25.23 & 24.02 & 23.94 & 24.15 & 23.33 \\
        \bottomrule
    \end{tabular}
    \caption{MAE for different text information categories.}
    \label{tab:mae}
\end{table}

\begin{table}[tpb]
    \centering
    \small
    \setlength{\tabcolsep}{2pt}
    \renewcommand{\arraystretch}{0.9}
    \begin{tabular}{lccccc}
        \toprule
        \textbf{Category} & \textbf{No\_text} & \textbf{Event\_info} & \textbf{Public} & \textbf{Traffic} & \textbf{TST} \\
        \midrule
        2km\_2h & 40.22 & 38.37 & 38.69 & 37.74 & 35.23 \\
        2km\_3h & 38.97 & 37.10 & 37.33 & 36.63 & 36.12 \\
        2km\_4h & 37.17 & 35.76 & 35.54 & 34.92 & 35.13 \\
        3km\_2h & 40.83 & 39.12 & 39.15 & 39.00 & 36.29 \\
        3km\_3h & 39.33 & 37.64 & 37.44 & 37.64 & 36.07 \\
        3km\_4h & 37.88 & 36.22 & 35.91 & 36.15 & 35.56 \\
        4km\_2h & 40.48 & 38.62 & 38.66 & 38.90 & 36.00 \\
        4km\_3h & 39.04 & 37.21 & 37.00 & 37.40 & 35.82 \\
        4km\_4h & 38.03 & 36.42 & 35.97 & 36.51 & 35.74 \\
        5km\_2h & 40.21 & 38.48 & 38.60 & 38.56 & 36.09 \\
        5km\_3h & 38.76 & 36.83 & 36.90 & 37.06 & 35.79 \\
        5km\_4h & 37.36 & 35.57 & 35.47 & 35.76 & 35.29 \\
        \bottomrule
    \end{tabular}
    \caption{RMSE for different text information categories.}
    \label{tab:rmse}
\end{table}

\section{Efficiency Analysis}
\label{App: Efficiency}
To evaluate the computational efficiency and practical applicability of our proposed model, we compare its inference time against several high-performing spatiotemporal methods, with results shown in Table~\ref{tab:inference_efficiency}.
\begin{table}[htbp] 
    \centering\small
    \setlength{\tabcolsep}{2pt}
    \renewcommand{\arraystretch}{0.9}
    \begin{tabular}{lcc}
    \toprule
    \textbf{Model} & \textbf{Inference Time (s)} & \textbf{Parameters} \\
    \midrule
    GWNET-GLA      & 0.0717    &  374K                \\
    GWNET-GBA      & 0.0259    &  344K                 \\
    DCRNN-GLA      & 0.2152    &  373K                  \\
    DCRNN-GBA      & 0.1395    &  373K                  \\
    DSTAGNN-GLA    & 0.2352    & 66.3M                  \\
    DSTAGNN-GBA    & 0.0906    & 26.9M                  \\
    Ours-GLA       & 0.2864    & \multirow{2}{*}{3.1M}                  \\
    Ours-GBA       & 0.2297    &                  \\
    \bottomrule
    \end{tabular}
    \caption{
Model Efficiency comparison. We compare our model with high-performance spatiotemporal models that capture both spatial and temporal dimensions. Baseline model times reflect inference for a large region at a single time point, whereas 'Ours' indicates inference per time step for affected sensors near a single venue. 'Parameters' is the number of learnable parameters. K: $10^3$, M: $10^6$. GLA: Greater Los Angeles Area, GBA: Greater Bay Area.}
    \label{tab:inference_efficiency} 
\end{table}

When examining these computational speeds, it is essential to acknowledge the distinct operational scopes: baseline spatiotemporal models typically generate predictions for an entire large-scale region at a single time point, whereas our model's reported inference time refers to processing data per time step for affected sensors specifically around a single venue. Despite handling multimodal inputs including pre-computed embeddings from broader regional spatiotemporal models along with event-specific textual features, our model demonstrates competitive inference speed. The observed processing time supports minute-level responsiveness, confirming its suitability for real-world dynamic mobility forecasting applications. It also operates with considerably fewer trainable parameters than the complex spatiotemporal model DSTAGNN. Furthermore, the data pre-processing stage is highly economical. The average one-time text analysis for a single event day using the Qwen3 API costs approximately \$0.016.

\begin{table*}[tpb]
    \centering
    \small
    \setlength{\tabcolsep}{4pt}
    \renewcommand{\arraystretch}{0.9}
    \begin{tabular}{lcccccccccccc}
        \toprule
        \multirow{3}[4]{*}{\textbf{Methods}} & \multicolumn{6}{c}{\textbf{By Time}} & \multicolumn{6}{c}{\textbf{By Type}} \\
        \cmidrule(lr){2-7} \cmidrule(lr){8-13} 
        & \multicolumn{2}{c}{\textbf{2h}} & \multicolumn{2}{c}{\textbf{3h}} & \multicolumn{2}{c}{\textbf{4h}} & \multicolumn{2}{c}{\textbf{2h}} & \multicolumn{2}{c}{\textbf{3h}} & \multicolumn{2}{c}{\textbf{4h}} \\
        \cmidrule(lr){2-3} \cmidrule(lr){4-5} \cmidrule(lr){6-7} \cmidrule(lr){8-9} \cmidrule(lr){10-11} \cmidrule(lr){12-13}
        & \textbf{MAE} & \textbf{RMSE} & \textbf{MAE} & \textbf{RMSE} & \textbf{MAE} & \textbf{RMSE} & \textbf{MAE} & \textbf{RMSE} & \textbf{MAE} & \textbf{RMSE} & \textbf{MAE} & \textbf{RMSE}\\
        \midrule
        LSTM & 28.77 & 39.53 & 26.49 & 37.82 & 25.92 & 37.84 & 30.24 & 41.76 & 27.83 & 39.92 & 27.20 & 39.89 \\
        PatchTST & 29.94 & 40.89 & 27.54 & 38.31 & 26.05 & 38.97 & 31.42 & 43.12 & 28.90 & 40.38 & 27.33 & 41.01 \\
        DCRNN & 28.83 & 39.45 & 27.46 & 39.25 & 27.52 & 39.08 & 30.73 & 42.23 & 29.01 & 42.05 & 28.67 & 41.13 \\
        AGCRN & 30.07 & 41.73 & 27.35 & 41.45 & 26.00 & 38.32 & 31.10 & 43.43 & 28.46 & 42.60 & 27.47 & 41.32 \\
        STGCN & 36.16 & 54.86 & 34.44 & 52.20 & 32.40 & 53.66 & 36.12 & 54.24 & 34.18 & 50.59 & 31.63 & 50.36 \\
        GWNET & 29.01 & 38.19 & 27.56 & 37.07 & 24.74 & 35.77 & 28.25 & 40.22 & 28.72 & 38.97 & 25.65 & 37.17 \\
        ASTGCN & 43.12 & 64.98 & 42.43 & 64.10 & 39.36 & 63.78 & 42.08 & 63.46 & 40.82 & 63.14 & 36.59 & 62.47 \\
        STGODE & 35.97 & 54.11 & 33.70 & 51.49 & 31.88 & 51.03 & 35.97 & 54.08 & 33.47 & 51.06 & 31.24 & 49.96 \\
        DSTAGNN & 29.45 & 40.33 & 28.62 & 39.99 & 28.14 & 39.97 & 31.05 & 43.20 & 29.90 & 42.50 & 28.86 & 41.94 \\
        \textbf{Ours} & \textbf{25.55} & \textbf{35.37} & \textbf{25.57} & \textbf{35.05} & \textbf{24.04} & \textbf{35.19} & \textbf{25.49} & \textbf{35.23} & \textbf{25.32} & \textbf{36.12} & \textbf{23.97} & \textbf{35.13} \\
        \bottomrule
    \end{tabular}
    \caption{Performance comparison of baseline models using sensor data within 2 km of event locations.}
    \label{tab:2km}
\end{table*}

\begin{table*}[tpb]
    \centering
    \small
    \setlength{\tabcolsep}{4pt}
    \renewcommand{\arraystretch}{0.9}
    \begin{tabular}{lcccccccccccc}
        \toprule
        \multirow{3}[4]{*}{\textbf{Methods}} & \multicolumn{6}{c}{\textbf{By Time}} & \multicolumn{6}{c}{\textbf{By Type}} \\
        \cmidrule(lr){2-7} \cmidrule(lr){8-13} 
        & \multicolumn{2}{c}{\textbf{2h}} & \multicolumn{2}{c}{\textbf{3h}} & \multicolumn{2}{c}{\textbf{4h}} & \multicolumn{2}{c}{\textbf{2h}} & \multicolumn{2}{c}{\textbf{3h}} & \multicolumn{2}{c}{\textbf{4h}} \\
        \cmidrule(lr){2-3} \cmidrule(lr){4-5} \cmidrule(lr){6-7} \cmidrule(lr){8-9} \cmidrule(lr){10-11} \cmidrule(lr){12-13}
        & \textbf{MAE} & \textbf{RMSE} & \textbf{MAE} & \textbf{RMSE} & \textbf{MAE} & \textbf{RMSE} & \textbf{MAE} & \textbf{RMSE} & \textbf{MAE} & \textbf{RMSE} & \textbf{MAE} & \textbf{RMSE}\\
        \midrule
        LSTM & 29.61 & 41.22 & 27.36 & 39.98 & 26.87 & 39.44 & 29.98 & 42.23 & 27.36 & 39.98 & 26.92 & 40.13 \\
        PatchTST & 29.58 & 41.14 & 27.05 & 39.65 & 26.59 & 39.22 & 29.94 & 42.15 & 27.05 & 39.65 & 26.64 & 39.91 \\
        DCRNN & 29.81 & 41.12 & 28.95 & 40.99 & 28.52 & 40.91 & 30.48 & 42.50 & 28.95 & 40.99 & 28.62 & 41.64 \\
        AGCRN & 32.19 & 47.05 & 30.49 & 48.85 & 30.21 & 51.65 & 31.23 & 45.48 & 30.49 & 48.85 & 28.51 & 47.04 \\
        STGCN & 35.53 & 52.30 & 33.16 & 51.42 & 32.42 & 51.04 & 34.93 & 51.49 & 33.16 & 51.42 & 30.91 & 48.07 \\
        GWNET & 28.93 & 39.55 & 26.96 & 39.04 & 25.58 & 37.97 & 28.93 & 40.48 & 27.36 & 39.04 & 25.58 & 38.03 \\
        ASTGCN & 42.09 & 59.46 & 40.35 & 57.29 & 39.37 & 56.16 & 40.59 & 57.91 & 40.35 & 57.29 & 35.86 & 53.16 \\
        STGODE & 35.34 & 51.47 & 32.94 & 49.37 & 31.79 & 48.32 & 34.78 & 51.32 & 32.94 & 49.37 & 30.52 & 47.59 \\
        DSTAGNN & 30.42 & 41.90 & 29.59 & 41.78 & 29.24 & 41.73 & 30.79 & 43.47 & 29.59 & 41.78 & 28.82 & 42.43 \\
        \textbf{Ours} & \textbf{25.08} & \textbf{36.38} & \textbf{24.73} & \textbf{36.80} & \textbf{23.79} & \textbf{36.42} & \textbf{24.93} & \textbf{36.00} & \textbf{24.22} & \textbf{35.82} & \textbf{23.69} & \textbf{35.74} \\
        \bottomrule
    \end{tabular}
    \caption{Performance comparison of baseline models using sensor data within 4 km of event locations.}
    \label{tab:4km}
\end{table*}

\begin{table*}[tpb]
    \centering
    \small
    \setlength{\tabcolsep}{4pt}
    \renewcommand{\arraystretch}{0.9}
    \begin{tabular}{lcccccccccccc}
        \toprule
       \multirow{3}[4]{*}{\textbf{Methods}} & \multicolumn{6}{c}{\textbf{By Time}} & \multicolumn{6}{c}{\textbf{By Type}} \\
        \cmidrule(lr){2-7} \cmidrule(lr){8-13} 
        & \multicolumn{2}{c}{\textbf{2h}} & \multicolumn{2}{c}{\textbf{3h}} & \multicolumn{2}{c}{\textbf{4h}} & \multicolumn{2}{c}{\textbf{2h}} & \multicolumn{2}{c}{\textbf{3h}} & \multicolumn{2}{c}{\textbf{4h}} \\
        \cmidrule(lr){2-3} \cmidrule(lr){4-5} \cmidrule(lr){6-7} \cmidrule(lr){8-9} \cmidrule(lr){10-11} \cmidrule(lr){12-13}
        & \textbf{MAE} & \textbf{RMSE} & \textbf{MAE} & \textbf{RMSE} & \textbf{MAE} & \textbf{RMSE} & \textbf{MAE} & \textbf{RMSE} & \textbf{MAE} & \textbf{RMSE} & \textbf{MAE} & \textbf{RMSE}\\
        \midrule
        LSTM & 28.40 & 39.85 & 26.41 & 38.84 & 25.98 & 38.40 & 29.06 & 41.23 & 26.75 & 39.84 & 26.24 & 39.24 \\
        PatchTST & 28.37 & 39.77 & 26.03 & 38.52 & 25.60 & 38.16 & 29.03 & 41.14 & 26.44 & 39.57 & 25.97 & 39.11 \\
        DCRNN & 28.81 & 39.97 & 28.12 & 39.98 & 27.77 & 39.99 & 29.81 & 41.74 & 28.64 & 41.15 & 28.04 & 40.83 \\
        AGCRN & 30.62 & 44.59 & 28.64 & 45.72 & 28.32 & 47.48 & 30.10 & 43.73 & 27.71 & 43.99 & 27.32 & 44.39 \\
        STGCN & 34.99 & 51.89 & 32.27 & 50.05 & 31.41 & 49.25 & 34.66 & 51.29 & 31.49 & 48.49 & 30.48 & 47.27 \\
        GWNET & 28.02 & 39.25 & 26.29 & 38.03 & 25.12 & 36.84 & 28.39 & 40.21 & 26.50 & 38.76 & 25.23 & 37.36 \\
        ASTGCN & 41.10 & 58.55 & 39.04 & 55.57 & 37.88 & 54.01 & 39.93 & 57.26 & 36.83 & 53.81 & 35.09 & 52.01 \\
        STGODE & 34.81 & 51.06 & 32.09 & 48.04 & 30.78 & 46.52 & 34.51 & 51.13 & 31.53 & 48.24 & 30.09 & 46.79 \\
        DSTAGNN & 29.41 & 40.72 & 28.74 & 40.74 & 28.46 & 40.78 & 30.12 & 42.68 & 28.78 & 41.90 & 28.23 & 41.59 \\
        \textbf{Ours} & \textbf{24.40} & \textbf{35.74} & \textbf{23.37} & \textbf{35.55} & \textbf{23.34} & \textbf{35.08} & \textbf{24.27} & \textbf{36.09} & \textbf{24.07} & \textbf{35.79} & \textbf{23.33} & \textbf{35.29} \\

        \bottomrule
    \end{tabular}
    \caption{Performance comparison of baseline models using sensor data within 5 km of event locations.}
    \label{tab:5km}
\end{table*}

\section{Pre-defined Spatiotemporal Keywords}
Table~\ref{tab:keywords} lists the pre-defined spatial and temporal keywords employed in our quantitative analysis in \textbf{RQ4}.  
\begin{table}[tbp]
\centering
\footnotesize 
\begin{tabular}{@{}lp{0.35\textwidth}@{}} 
\toprule
\textbf{Category} & \textbf{Keywords} \\
\midrule
\textbf{Spatial} & Location, Venue, Arena, Stadium, Street, Road, Avenue, Highway, Intersection, District, Zone, Area, Region, Downtown, Map, Route, Address, Coordinates, Near, Vicinity, Surrounding, Adjacent, Within, Across, Along, Between, Entrance, Exit, Parking, Radius \\
\addlinespace 
\textbf{Temporal} & Hour, Minute, Day, Week, Month, Morning, Afternoon, Evening, Night, AM, PM, Clock, Today, Date, Schedule, Timeline, Duration, Period, Before, After, Early, Late, Start, End, During, Arrival, Departure, Peak, Weekend, Daily \\
\bottomrule
\end{tabular}
\caption{Pre-defined Spatial and Temporal Keywords.}
\label{tab:keywords}
\end{table}

\section{Main Prompts}
\label{APP:prompt}
\subsection{Prompts for Event Info Extractor}
\label{APP:prompt_event}
After acquiring metadata from the Venue Calendar Database for the event day, the prompt given to the Event Info Extractor is as follows:
\begin{tcolorbox}[
    colback=gray!15!white,
    colframe=gray!60!black,
    arc=1mm,
    boxrule=0.5pt,
    width=\columnwidth,
    enlarge left by=0mm,
    enlarge right by=0mm,
    left=2mm,
    right=2mm,
    top=3mm,
    bottom=3mm,
    breakable 
]
\noindent \textbf{Objective:} Please analyze the following event information and extract the key details in a clear text format.

\medskip

\noindent \textbf{Required Output Structure:} Please extract and organize the following information in a clear text format (using titles to separate each section):
\begin{enumerate}[
    label=\arabic*., 
    nosep, 
    leftmargin=*, 
    rightmargin=0pt 
]
    \item \textbf{Event Type:} Clearly define the category of the event (such as sports event, concert, music festival, exhibition, public activity, etc.) and briefly explain its core features or definition to help identify similar event types.
    \item \textbf{Event Venue and Location Information:} Provide the name of the venue and its significant features (such as capacity, facilities, technical equipment, geographical advantages or limitations), and describe the characteristics of the surrounding area (such as transportation convenience, commercial districts, residential areas, natural landscapes, etc.), to showcase how the venue may impact the event.
    \item \textbf{Event Time:} Provide the start time and estimated duration; if the official duration is not provided, reasonably estimate it based on the typical duration for this type of event, and explain the basis of your assumption.
    \item \textbf{Event Content:} Summarize the main activities, goals, and unique highlights of the event, including key people, teams, or organizations involved, their background, fame, or influence (such as international stars, local celebrities, authoritative organizations, etc.). If there are special segments (such as fan meet-ups, opening ceremonies, etc.), mention them.
    \item \textbf{Target Audience:} Describe the characteristics of potential participants, including age range, interests (such as music lovers, technology enthusiasts), or professional background (such as students, professionals), and analyze their motivation for attending (such as entertainment, learning), along with the proportion of the audience.
    \item \textbf{Event Scale or Importance:} Specify if this is a "locally focused event," "regionally influential activity," or "nationally appealing event," or estimate the number of participants based on the venue’s capacity and event type (such as "about hundreds of participants," or "estimated thousands in attendance"). When estimating, consider the venue's maximum capacity as the upper limit, the event's appeal (e.g., regular activities tend to attract fewer attendees compared to well-known artists or major championships), and the reasonableness of the estimate.
\end{enumerate}
Please think step-by-step.
\medskip

\noindent \textbf{Output:} \textit{<basic event information>}
\end{tcolorbox}
\subsection{Prompts for Tweet Analyzer}
\label{APP:prompt_tweet}
Using the event information provided by the Event Info Extractor, the agent constructs retrieval keywords by adhering to the following prompt:
\begin{tcolorbox}[
    colback=gray!15!white,  
    colframe=gray!60!black, 
    arc=1mm,                
    boxrule=0.5pt,          
    width=\columnwidth,     
    enlarge left by=0mm,    
    enlarge right by=0mm,   
    left=3mm,               
    right=3mm,
    top=3mm,
    bottom=3mm,
    breakable              
]
Please use the \textit{<basic event information>} to create a set of targeted Twitter search queries. Your goal is to generate 5 distinct, logically constructed search query strings. Each query string must strictly adhere to the specific structure:
\begin{center} 
\texttt{(EventRelatedTermA OR EventRelatedTermB OR ...)} \\
\texttt{AND} \\
\texttt{(LocationRelatedTermX OR LocationRelatedTermY OR ...)}
\end{center}
To populate this structure:
\begin{itemize}[nosep, leftmargin=*, itemsep=2pt, topsep=3pt] %
    \item The \textbf{'Event-Related' component} (terms joined by \texttt{OR}) should relate to the event's name, any common variations or nicknames, keywords representing key activities or the primary event type, and relevant event-specific hashtags.
    \item The \textbf{'Location-Related' component} (terms joined by \texttt{OR}) should relate to the venue name or nicknames, the city, and optionally, other crucial official location identifiers (like a distinct district or campus name if provided and applicable) to maximize specificity.
\end{itemize}
Please think step-by-step.
\end{tcolorbox}
For each event, tweets are initially retrieved using five constructed queries and then deduplicated. The retrieved tweets are subsequently analyzed by the agent, guided by the following prompt:

\begin{tcolorbox}[
    colback=gray!15!white,  
    colframe=gray!60!black, 
    arc=1mm,                
    boxrule=0.5pt,          
    width=\columnwidth,     
    enlarge left by=0mm,    
    enlarge right by=0mm,   
    left=3mm,               
    right=3mm,
    top=3mm,
    bottom=3mm,
    breakable              
]
Please filter the tweets related to \textit{<basic event information>} and analyze the following batch of tweets with a focus on the following aspects and provide a concise, structured response:
\begin{enumerate}[label=\arabic*., nosep, leftmargin=*, itemsep=3pt, topsep=3pt] 
    \item \textbf{Social Media Attention and Reasons:} Evaluate the attention trend and performance of the event on social media, using descriptive language (e.g., "widely discussed," "moderate attention," or "limited attention"), and explain the reasons (e.g., topic appeal, dissemination range, or time factors).
    \item \textbf{Public Participation Willingness and Audience Characteristics with Reasons:} Describe the strength of public willingness to attend the event (e.g., "strong willingness to participate," "some groups are interested," or "willingness to participate is unclear") and the characteristics of potential participants (e.g., age, interests, or professional groups), and provide reasoning (e.g., the nature of the event, convenience, or alignment with target audiences).
    \item \textbf{Sentiment Distribution and Reasons:} Summarize the sentiment tendencies on social media related to the event (e.g., "generally excited," "somewhat positive," "neutral," or "negative emotions dominate") and analyze the reasons (e.g., event highlights, controversy, or public expectations).
    \item \textbf{Main Discussion Topics:} Extract and list the main topics or keywords related to the event in tweets or social media discussions (e.g., event content, key individuals, or points of controversy), keeping it brief.
\end{enumerate}

\medskip 

\noindent \textbf{Output:} \textit{<social media analysis results>}
\end{tcolorbox}
\subsection{Prompts for Mobility Analyzer}
This agent processes information from the first two agents to conduct spatiotemporal text filtering pertinent to traffic analysis, according to the prompt below:
\begin{tcolorbox}[
    colback=gray!15!white,  
    colframe=gray!60!black, 
    arc=1mm,                
    boxrule=0.5pt,          
    width=\columnwidth,     
    enlarge left by=0mm,    
    enlarge right by=0mm,   
    left=3mm,               
    right=3mm,
    top=3mm,
    bottom=3mm,
    breakable              
]
\noindent\textbf{Role:} You are a premier expert in California traffic impact prediction, possessing extensive professional experience and a profound understanding of the state's diverse cultural fabric, unique urban road networks, and varied residential patterns.

\medskip
\noindent\textbf{Objective:} Conduct a preliminary analysis of an event's impact on surrounding road traffic for a specific day.

\medskip
\noindent\textbf{Methodology:} Please think step-by-step.
\begin{enumerate}[label=\arabic*., itemsep=1pt, topsep=2pt, leftmargin=*]
    \item First, synthesize the \textit{<basic event information>}, relevant data from \textit{<recent events>}, and insights from \textit{<social media analysis results>}.
    \item Second, utilize the specified \textit{<logic>} to structure your assessment.
    \item Third, your analysis must explicitly consider and apply your expert knowledge of California's unique urban characteristics and residential behaviors.
\end{enumerate}

\medskip
\noindent\textbf{Required Output Structure:} Your report detailing the preliminary analysis must be formatted precisely as follows. Focus exclusively on road traffic.

\medskip\noindent\textbf{\#Traffic conditions}\par\smallskip
\begin{itemize}[label=\textbullet, itemsep=1pt, topsep=2pt, leftmargin=*]
    \item \textbf{General Trend of Traffic Flow Changes:} Describe anticipated shifts in traffic volume (e.g., percentage increase), congestion levels (e.g., severe, moderate, light), and specific roadways likely to be affected.
    \item \textbf{Impact Range:} Estimate the geographical extent of traffic effects (e.g., radius in miles/km from the event, specific intersections, affected freeway segments).
    \item \textbf{Duration:} Predict the timeframe of the traffic impact, including estimated start, peak congestion, and when traffic is expected to normalize.
    \item \textbf{Detailed Reasoning:} Provide a thorough step-by-step explanation for your conclusions regarding the traffic conditions. Directly link your predictions back to the \textbf{\#Filtered event info}, \textbf{\#Filtered public reactions}, the specified \textit{<logic>}, and your expert knowledge of California's traffic dynamics and cultural patterns. Explain why these factors lead to the predicted outcomes.
\end{itemize}

\medskip\noindent\textbf{\#Filtered event info}\par\smallskip
Present the key elements from the \textit{<basic event information>} and \textit{<recent events>} that were most influential in your traffic impact assessment. Highlight specific details (e.g., precise location, timing relative to peak hours, scale of event, relevant comparisons to past events).

\medskip\noindent\textbf{\#Filtered public reactions}\par\smallskip
Summarize the salient points from the \textit{<social media analysis results>} that significantly shaped your predictions. Focus on aspects indicating potential crowd size beyond official estimates, geographic origin of attendees, and overall public intent to travel to the event area.

\medskip
\noindent The output should be in JSON formats: \texttt{\{"filtered event info": "...", "filtered public reactions": "...", "traffic conditions": "..."\}}

\medskip
\noindent\textbf{Output:} \textit{<filtered event info>}, \textit{<filtered public reactions>}, \textit{<traffic conditions>}
\end{tcolorbox}
\subsection{Prompts for Evaluator}
The <$logic$> component consists of two parts: \textit{<$logic\_global$>}, which outlines general screening principles, and \textit{<$logic\_venue$>}, which provides venue-specific guidelines. The prompt for the Evaluator agent to revise \textit{<$logic\_global$>} is as follows:
\begin{tcolorbox}[
    colback=gray!15!white,  
    colframe=gray!60!black, 
    arc=1mm,                
    boxrule=0.5pt,          
    width=\columnwidth,     
    enlarge left by=0mm,    
    enlarge right by=0mm,   
    left=3mm,               
    right=3mm,
    top=3mm,
    bottom=3mm,
    breakable              
]
\noindent\textbf{Objective:} Analyze aggregated historical prediction error statistics to identify systemic weaknesses and patterns in the general information screening logic. Propose updates to enhance its overall accuracy and robustness across diverse events and venues.

\medskip
\noindent\textbf{Input Data:}
\begin{itemize}[label={}, leftmargin=0.5em, itemsep=1pt, topsep=2pt, parsep=0pt] 
    \item \textit{<Error\_Patterns\_By\_Day\_Of\_Week>}: Statistical breakdown of errors by weekday.
    \item \textit{<Error\_Patterns\_By\_Time\_Of\_Day>}: Statistical breakdown of errors by specific time slots.
    \item \textit{<Error\_Patterns\_By\_Event\_Type>}: Statistics on which types of events most frequently exhibit high prediction errors.
\end{itemize}

\medskip
\noindent\textbf{Instructions:} Identify systemic patterns that indicate deficiencies in our current general information screening \textit{<logic\_global>}. Your goal is to understand why these broad error trends are occurring and how to refine the logic that filters and weighs information. Consider:
What common characteristics link the days, times, or event types with the highest error rates? What types of information might be particularly relevant or deceptive in these high-error contexts that our general logic isn't capturing well?
Please think step-by-step.
\medskip

\noindent\textbf{Output:} Updated \textit{<logic\_global>}
\end{tcolorbox}
The prompt for the Evaluator agent to revise \textit{<$logic\_venue$>} is as follows:
\begin{tcolorbox}[
    colback=gray!15!white,  
    colframe=gray!60!black, 
    arc=1mm,                
    boxrule=0.5pt,          
    width=\columnwidth,     
    enlarge left by=0mm,    
    enlarge right by=0mm,   
    left=3mm,               
    right=3mm,
    top=3mm,
    bottom=3mm,
    breakable              
]
\noindent\textbf{Objective:} Analyze aggregated historical prediction error statistics for a specific venue to identify unique error patterns. Propose targeted updates to create or refine a venue-specific information screening logic.

\medskip
\noindent\textbf{Input Data:}
\begin{itemize}[label={}, leftmargin=0.5em, itemsep=1pt, topsep=2pt, parsep=0pt] 
    \item \textit{<Venue\_Name>}: The specific venue being analyzed.
    \item \textit{<High\_Error\_Event\_Profile\_At\_Venue>}: A statistical summary of events at this venue that had the most severe prediction errors. This includes: Common event types/scales that consistently result in high errors at this venue. Frequently problematic time periods. Specific locations around the venue that are repeatedly sites of significant prediction errors.
\end{itemize}

\medskip
\noindent\textbf{Instructions:} Identify error patterns of this venue. Your goal is to determine how information screening logic of this venue \textit{<logic\_venue>} should be refined, considering its unique \textit{<Venue\_Profile>}: What common threads link the high-error events, times, and locations detailed in the \textit{<High\_Error\_Event\_Profile\_At\_Venue>}? How might these relate to the venue's physical characteristics, typical event portfolio, or surrounding environment?
Please think step-by-step.

\medskip
\noindent\textbf{Output:} Updated \textit{<logic\_venue>}
\end{tcolorbox}
The following sample demonstrates the revision of the screening logic for improved real-world accuracy:
\begin{tcolorbox}[
    colback=gray!15!white,  
    colframe=gray!60!black, 
    arc=1mm,                
    boxrule=0.5pt,          
    width=\columnwidth,     
    enlarge left by=0mm,    
    enlarge right by=0mm,   
    left=3mm,               
    right=3mm,
    top=3mm,
    bottom=3mm,
    breakable              
]
Here is the analysis of prediction errors for the Hollywood Bowl:
\begin{itemize}[nosep, leftmargin=*, itemsep=2pt, topsep=3pt]
    \item \textbf{Insufficient Pre-Event Window for Popular Concerts:} Error curves clearly indicate that the standard 1.5-hour pre-event window for anticipating traffic build-up is inadequate for popular weekend concerts. This is due to attendees typically arriving much earlier, often for pre-event activities like picnicking. The system's logic needs to be triggered by a combination of event category (e.g., pop concert), day of the week (weekend), and indicators of high public interest (e.g., expected attendance ratio, social media buzz).
    \item \textbf{Inadequate Egress Time for Sold-Out Events:} For sold-out concerts, especially considering the venue's congested parking and limited exit routes, the previously allocated 45-minute post-event window for traffic normalization is consistently proving insufficient.
    \item \textbf{Underestimation of Congestion on Key Routes:} Specific major access routes, namely Highland Avenue (for ingress) and Cahuenga Boulevard East (for both ingress and egress), are persistently underestimated in terms of congestion severity and duration during these identified high-impact events. The screening logic must more emphatically flag these critical segments.
\end{itemize}

\medskip
\noindent Based on these principles, the information screening logic for this venue has been updated as follows:

\medskip
\noindent\textbf{Updated Logic Rules:}

\medskip
\noindent\textbf{Adjusted Ingress Window for High-Demand Weekend Concerts:}\\
For events identified as popular pop concerts occurring on Fridays, Saturdays, or Sundays with high anticipated attendance, the system will now recognize the onset of significant ingress traffic impact starting two hours prior to the official event commencement. Furthermore, predicted traffic volumes within the 1.5-hour window immediately preceding the event start time will be considered substantially more intense than under previous calculations.

\medskip
\noindent\textbf{Extended Egress Impact Period for High-Demand Weekend Concerts:}\\
For the same category of popular weekend pop concerts with high anticipated attendance, the duration of significant post-event traffic impact is now projected to persist for at least one hour following the event's conclusion. The initial thirty minutes of this period, in particular, will be recognized as having an intensified level of congestion and slower dispersion rates.

\medskip
\noindent\textbf{Enhanced Flagging for Critical Road Segments During High-Impact Scenarios:}\\
When conditions indicative of a 'Popular Weekend Pop Concert with High Anticipated Attendance' are met (as per the updated ingress and egress timing logic), the screening logic will now apply special high-alert designators to specific, historically problematic road segments. Notably, Highland Avenue will be flagged for heightened early ingress congestion, and Cahuenga Boulevard East will be flagged for severe and prolonged congestion during both ingress and egress phases. This ensures these critical arteries receive priority attention in subsequent traffic impact assessments.
\end{tcolorbox}
\end{document}